\documentclass{amia}
\usepackage{lipsum} 
\usepackage{graphicx}
\usepackage{xcolor}
\usepackage{booktabs}
\usepackage{multirow}
\usepackage{wrapfig,booktabs}
\usepackage{amsmath}
\definecolor{darker}{rgb}{0.15,0.45,0.4}
\usepackage[colorlinks, urlcolor=darker, citecolor=darker, linkcolor=darker]{hyperref} 

\usepackage[table]{xcolor}
\definecolor{pastelgreen}{HTML}{E0F4DA}
\definecolor{headgray}{gray}{0.9}        % light gray header
\definecolor{pastelgreen}{HTML}{E0F4DA} 
\usepackage[normalem]{ulem}
\usepackage{xcolor}
\definecolor{blueKey}{HTML}{2B6CB0}
\definecolor{greenKey}{HTML}{2F855A}
\definecolor{redKey}{HTML}{C53030}

\begin{document}

\newif\ifanonymous
 \anonymoustrue      % <- use this for submission (anonymous)

\ifanonymous

\title{Knowledge Graph Augmented Large Language Models for Disease Prediction}

\author{Anonymous Authors}
\institutes{Affiliations withheld for double-blind review}

\title{Knowledge Graph Augmented Large Language Models for Disease Prediction}

\author{
  Ruiyu Wang$^1$,
  Tuan Vinh$^{2}$,
  Ran Xu$^1$,
  Yuyin Zhou, PhD$^3$,
  Jiaying Lu, PhD$^4$, 
  Francisco Pasquel, MD$^5$, 
  Mohammed K Ali, MD$^6$,
  Carl Yang, PhD$^{1, 4, 6}$
}

\institutes{
  $^1$ Department of Computer Science, Emory University, Atlanta, GA, USA \\
  $^2$ Division of Medical Sciences, Oxford University, Oxford, UK \\
  $^3$ Department of Computer Science and Engineering, UCSC, CA, USA \\
  $^4$ Nell Hodgson Woodruff School of Nursing, Emory University, Atlanta, GA, USA \\
  $^5$ School of Medicine, Emory University, Atlanta, GA, USA \\
  $^6$ Rollins School of Public Health, Emory University, Atlanta, GA, USA \\
}

\fi

% -human evaluation
% -collaborator and student track
% - baseline for cradle
% - overall how to emphasize the contribution
% -appendix
\maketitle

\section*{Abstract}
\label{sec:abstract}
Electronic health records (EHRs) support strong clinical prediction but often provide coarse, post hoc explanations that are hard to use for patient-level decisions. We propose a knowledge-graph (KG)–guided chain-of-thought (CoT) framework for visit-level disease prediction on MIMIC-III. We map ICD-9 codes to PrimeKG, mine disease-relevant nodes and paths, and use these paths to scaffold temporally consistent CoT explanations, retaining only samples whose conclusions match observed outcomes. We then fine-tune lightweight LLaMA-3.1-Instruct-8B and Gemma-7B models on two small cohorts (400 and 1,000 index visits) across ten PrimeKG-mapped diseases. Our models outperform strong classical baselines, reaching AUROC of 0.66–0.70 and macro-AUPR of 0.40–0.47. Without additional training, the models transfer zero-shot to the CRADLE cohort, improving accuracy from 0.40–0.51 to 0.72–0.77. Blinded clinicians consistently prefer KG-guided CoT for clarity, relevance, and correctness. Code is available at \texttt{https://github.com/JonathanWry/KG-guided-LLM-pipeline}.

\begin{figure}[h] 
  \centering
  \vspace{-0.1cm}
  \includegraphics[width=.7\linewidth]{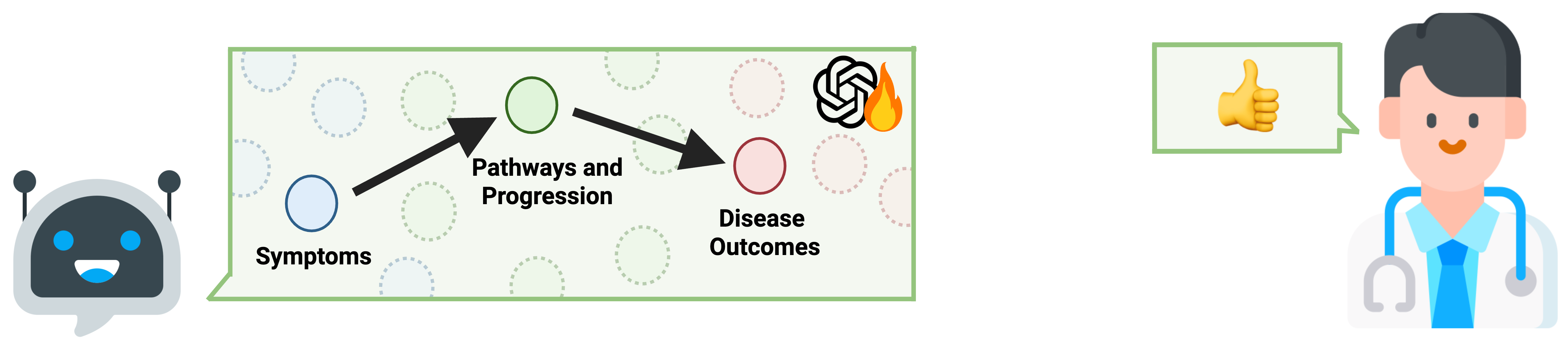}
  \label{fig:my_figure}
\end{figure}
\vspace{-0.1cm}
\section*{Introduction}
\label{sec:intro}
Electronic health records (EHRs) are now routinely collected across large and diverse healthcare systems, capturing longitudinal data on diagnoses, medications, and laboratory results from millions of patients.\cite{king2014clinical,ruan2019representation,zheng2014automatically} As interest in digital medicine grows,\cite{fogel2018ai,landi2020deep} these data are increasingly used to train predictive models for outcomes such as in-hospital mortality, readmission, and disease onset, often using benchmark cohorts like MIMIC-III.\cite{johnson2016mimic,goldstein2017practical} Classical models such as logistic regression and tree-based ensembles remain popular due to their robustness, global interpretability, and ease of deployment in clinical systems,\cite{goldstein2017practical} while neural architectures including recurrent and attention-based models have been explored to capture temporal dynamics and co-occurrence patterns in EHR data.\cite{choi2016retain,ruan2019representation} Despite these advances in predictive performance, most EHR models provide only coarse, post hoc interpretability (such as global feature importance, attribution scores, or attention weights), offering limited support for patient-specific, clinician-facing reasoning about why a particular outcome is likely.\cite{goldstein2017practical,choi2016retain,caruana2015intelligible,rudin2019stop}

Advances in large language models (LLMs) and reasoning-oriented architectures offer a potential remedy: LLMs can generate intermediate reasoning steps alongside predictions, enabling explicit chain-of-thought (CoT) narratives that articulate why a given prediction should hold.\cite{wei2022chain,wang2024a,huang202202cot,xie2024reasoning} In principle, such step-by-step explanations could bridge the gap between accurate prediction and clinically usable justification for clinical predictions. However, applying CoT in medicine remains challenging. High-quality, clinically validated CoT datasets are scarce \cite{muennighoff2025,ye2025} and unconstrained LLM reasoning can introduce hallucinated or medically incorrect intermediate steps, which is particularly problematic in high-stakes clinical decision making.\cite{xu2024ram,liu2025ragreview} These limitations have motivated efforts to constrain or ground LLM reasoning using external knowledge sources to improve factual reliability. Biomedical knowledge graphs (KGs) such as PrimeKG integrate curated relationships among diseases, phenotypes, drugs, and genes, and have been leveraged to enhance factual grounding in medical question answering and decision-support systems.\cite{chandak2023primekg,liu2025ragreview}

In this paper, we connect these previous works by integrating biomedical KG structure directly into the CoT generation process for visit-level disease prediction on structured EHR data. We propose a KG-guided CoT framework for MIMIC-III\cite{johnson2016mimic} that produces clinically grounded reasoning rather than unconstrained free-form explanations. In our formulation, each visit-level instance uses ICD-9 code features from an index visit $t$ to predict a binary disease label at the immediately subsequent visit $t{+}1$. Concretely, we map ICD-9 codes to PrimeKG\cite{chandak2023primekg} through a three-stage entity-alignment procedure combining exact matching, embedding-based similarity retrieval, and LLM-driven clinical validation. Using these mapped entities, we identify disease-specific relevant features and extract KG paths linking each feature to its corresponding disease, pruning spurious or clinically irrelevant chains with GPT-4o. Guided by these paths, we prompt an LLM to generate temporally consistent, KG-anchored CoT explanations of whether a disease will appear at the next visit, and we retain only those reasoning traces whose conclusions match the ground-truth label. We then fine-tune lightweight open-weight models on this KG-anchored CoT supervision corpus so that, at inference time, they take ICD-9 features together with disease-specific KG evidence as input and produce both predictions and clinician-style explanations. This enables data-efficient, interpretable patient-level prognostic reasoning grounded in biomedical knowledge graphs.

We evaluate this framework on ten PrimeKG-mapped diseases in MIMIC-III across small patient cohorts (400 and 1{,}000 patient index visits), where our KG-anchored CoT supervision enables LLMs to outperform strong classical baselines such as XGBoost and Random Forest, achieving AUROC of approximately 0.67 and 0.70 and macro-AUPR of approximately 0.40 and 0.47, respectively. Clinician evaluations further show that KG-guided CoT yields substantially clearer, more coherent, and more clinically sound reasoning across three dimensions: clarity \& coherence, coverage \& relevance, and correctness \& soundness. Finally, our CoT-tuned models transfer effectively to the CRADLE cohort for forecasting cardiovascular events within one year after type 2 diabetes, improving accuracy from roughly 0.40–0.51 (untuned LLMs) to 0.72–0.77 without any retraining, and providing more temporally consistent and clinically plausible explanations. Together, these results demonstrate that KG-anchored CoT supervision enables small LLMs to learn data-efficient, clinically grounded prognostic reasoning that generalizes across cohorts.
\section*{Method}

\begin{figure}[h] 
\vspace{-0.5cm}
  \centering
  \includegraphics[width=1\linewidth]{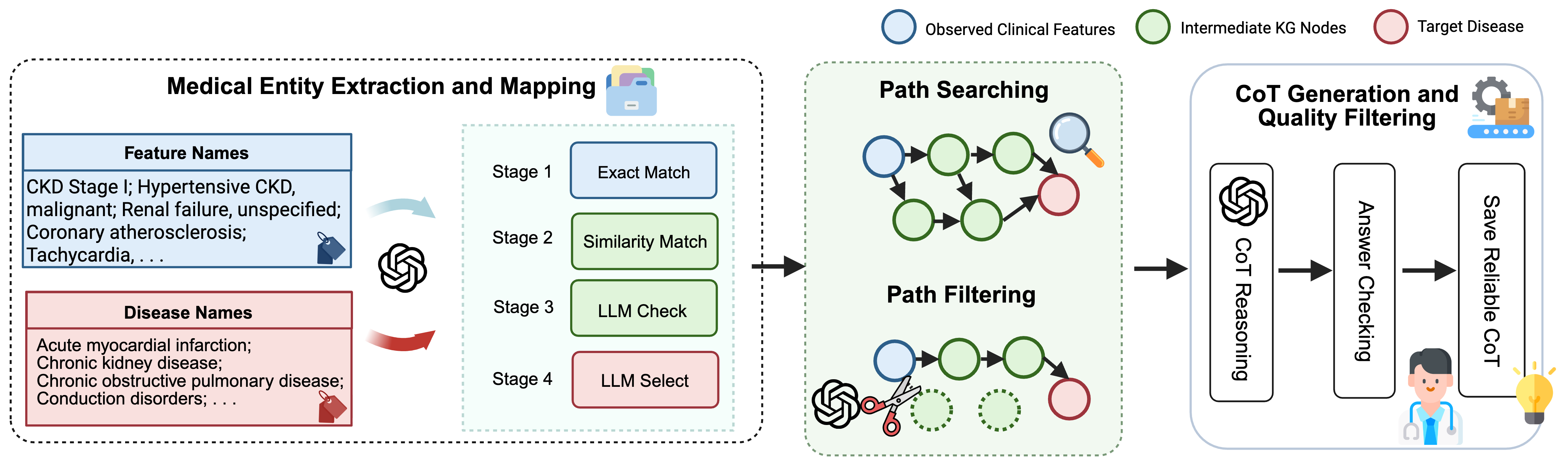}
  \vspace{-0.4cm}
  \caption{General pipeline for KG-guided CoT data generation. \textbf {Observed clinical features} are denoted in \textcolor{blueKey}{Blue}; 
\textbf{Intermediate KG nodes} are denoted in \textcolor{greenKey}{Green},
and the \textbf{Target disease} is denoted in \textcolor{redKey}{Red}.}
  \label{fig:my_figure}
\end{figure}

\paragraph{Preliminaries.}
A knowledge graph (KG) is a structured representation of entities and their relationships, typically modeled as a graph $G = (V, E)$ where $V$ denotes the set of entities (nodes) and $E$ the set of typed edges encoding relations among them. In the biomedical setting, KGs organize curated knowledge about diseases, phenotypes, drugs, genes, and related concepts, and provide an explicit relational backbone that can be used to guide or constrain model reasoning. Since biomedical relations in our external knowledge source are represented as direction- and type-specific (e.g., disease $\rightarrow$ gene and gene $\rightarrow$ disease may be encoded as distinct edges), we treat the underlying biomedical KG as a heterogeneous \emph{directed} multigraph. Specifically, we adopt PrimeKG\cite{chandak2023primekg} as our external knowledge source for grounding visit-level prognostic reasoning. PrimeKG is a large-scale biomedical KG that integrates curated relationships among diseases, phenotypes, drugs, genes, and other biomedical entities into a unified graph. We use it as the substrate for (1) aligning structured EHR concepts to KG nodes, (2) identifying disease-relevant neighborhood regions, and (3) extracting reasoning paths that connect clinical features to target diseases.

\paragraph{Prediction Task.}
We consider a next visit disease prediction task over structured EHR data.
For each patient, let $\{x_t\}_{t=1}^T$ denotes a sequence of visits ordered in time,
where each visit is represented as a binary feature vector $x_t \in \{0,1\}^{N}$, with $N$ the dimensionality of the code space. Let $\mathcal{D}$ denotes a set of disease labels of interest, and for each $d \in \mathcal{D}$ and visit index $t$, define a binary outcome $y_{t+1}^{(d)} \in \{0,1\}$, indicating whether disease $d$ is present at the next visit $t{+}1$. The model’s goal is, for each pair $(x_t, d)$, to estimate a probability $\hat{p}^{(d)}_{t+1} = f(x_t, d)$ and an associated binary decision $\hat{y}^{(d)}_{t+1}$ about whether $d$ will appear at visit $t{+}1$. Our KG-guided models are additionally conditioned on disease-specific KG evidence (relevance sets and reasoning paths) associated with $d$, as described below. In our experiments, we instantiate this general setup using ICD\textendash9 diagnosis codes as features, with $N = 7{,}423$ and $|\mathcal{D}| = 10$ diseases on the MIMIC-III cohort (details in the Results section).

% \paragraph{Using $x_t$ for next-visit prediction.}
% Following the next-visit setup of Harutyunyan et al.~\cite{harutyunyan2019multitask}, we condition prediction on the diagnosis codes observed at the index visit $t$ (i.e., $x_t$) to isolate the effect of KG-grounded CoT supervision under a widely used and easily reproducible formulation. 
% Although longer patient history $\{x_{t-m},\ldots,x_t\}$ can carry additional signal, $x_t$ already captures many persistent comorbidities and active problems recorded at the current encounter and matches our visit-pair construction protocol (Section~\ref{sec:results}). 
% Multi-visit context can be added by aggregating a lookback window without changing KG evidence construction or filtering.

\paragraph{KG Entity Mapping.}
To incorporate structured biomedical knowledge, we align ICD-9 concepts from MIMIC-III
to PrimeKG nodes through a three-stage mapping procedure. Let $E_{\text{ICD}}$ denote the
set of unique ICD-9 descriptions, and let $V_G$ be the node set of PrimeKG. For each entity
$e_i \in E_{\text{ICD}}$, we retrieve a similarity-ranked candidate set
$S_i = \{ s_1, \dots, s_C \} \subset V_G$ obtained by encoding $e_i$ and all KG node labels using a
text embedding model and ranking candidates by cosine similarity. The final mapped entity $\hat{e}_i$ is selected in three stages:

\textbf{Stage 1 (Exact match).}
If the ICD-9 text of $e_i$ exactly matches a node label $s_c$ in $S_i$, we record
$\tilde{e}_i = s_{c}$.

\textbf{Stage 2 (Similarity match).}
If an exact match is not found and the top similarity score exceeds a predefined
threshold $\tau$ (set to 0.85), we select the most similar entity:
\[
\tilde{e}_i = \arg\max_{s_c \in S_i} \cos(e_i, s_c)
\quad \text{if} \quad
\max_{s_{c} \in S_i} \cos(e_i, s_c) > \tau.
\]

\vspace{-0.2cm}
\textbf{Stage 3 (LLM-based filtering).}
All provisional mappings $\{ \tilde{e}_i \}_{i=1}^n$ from Stages~1 and~2 are then passed to
GPT-4o, which validates or revises them to ensure clinical correctness, yielding the final
mapped entities
\[
\hat{e}_i = \mathrm{LLM}(e_i, \tilde{e}_i \mid I_{\text{select}}),
\]
where $I_{\text{select}}$ is the prompt that asks the LLM to confirm, correct, or
reject the candidate mapping based on clinical and biomedical plausibility. This three-stage procedure produces a final set of 1{,}513 mapped ICD-9 feature
nodes and 10 mapped disease nodes, which serve as anchors for downstream relevance mining
and reasoning path construction.

\paragraph{KG Relevance Node and Path Mining.}
Let $\{\hat e_i^{f}\}_{i=1}^{n}$ denote the mapped ICD-9 feature nodes and
$\{\hat e_d\}_{d \in \mathcal{D}}$ denote the mapped disease nodes in PrimeKG
(here $n{=}1{,}513$ and $|\mathcal{D}|{=}10$). Each $\hat e_d$ corresponds to one of the
target diseases, while the $\hat e_i^{f}$ represent potential risk factors, comorbidities, or
intermediate biomedical concepts. Our goal is to derive, for each disease $d \in \mathcal{D}$, a compact set of KG
reasoning paths linking a small subset of feature nodes to that disease. For disease-centric node relevance, we present the disease node $\hat e_d$ and the set
of all mapped feature nodes $\{\hat e_i^{f}\}_{i=1}^{n}$ to GPT-4o and identify the
most relevant features for predicting the presence of $\hat e_d$ at $t{+}1$. This yields relevance set:
\[
\mathcal{R}_d = \{ \hat e_{k,d} \}_{k=1}^{K_{\text{node}}}
= \mathrm{LLM}(\{\hat e_i^{f}\}_{i=1}^{n}, \hat e_d \mid I_{\mathrm{node\_select}}),
\quad K_{\text{node}} = 8 \; ,
\]
where each $\hat e_{k,d}$ is a mapped feature node selected from $\{\hat e_i^{f}\}_{i=1}^{n}$ for disease $d$. Given $\mathcal{R}_d$, we extract reasoning chains from PrimeKG by computing all shortest paths between each feature node $\hat e_{k, d} \in \mathcal{R}_d$ and the disease node $\hat e_d$. Following prior work on KG-grounded reasoning \cite{luo2025o1pruner,chen2024overthinking}, we use shortest paths to limit ``overthinking’’ and preserve only the most immediate
biomedical relations:
\[
\tilde{\mathcal{P}}_{k,d} = \mathrm{shortest\_path}\!\big(\hat e_{k,d}, \hat e_d, G, L\big),
\]
\[
\mathcal{P}_d
= \{ P_{k,d} \}_{k=1}^{K_{\text{path}}}
= \mathrm{LLM}\!\left(
    \big\{ \tilde{\mathcal{P}}_{k,d} : \hat e_{k,d} \in \mathcal{R}_d \big\},
    \hat e_d \,\middle|\, I_{\mathrm{path\_select}}
\right),
\quad K_{\text{path}} = 5,
\]
where \(\mathrm{shortest\_path}(\cdot,\cdot; G, L)\) denotes the set of all minimum-length paths
between two nodes in the biomedical knowledge graph \(G\) (PrimeKG), subject to a maximum hop bound \(L\) (we use \(L{=}5\) in our experiments) to avoid overly long, weakly supported chains. Since multiple paths may exist for a disease, many of which are clinically irrelevant, weakly supported, or redundant, we apply GPT-4o under a path-selection instruction $I_{\mathrm{path\_select}}$ to obtain a disease-level path set, yielding a compact subgraph capturing mechanistic or epidemiologic links from relevant clinical concepts to disease $d$. These subgraphs serve as the KG-grounded evidence used in our subsequent CoT generation.

% \paragraph{Path pruning: correctness, stability, and coverage.}
% The shortest-path enumeration $\{\tilde{\mathcal{P}}_{k,d}\}$ provides a conservative candidate pool: every candidate is composed of existing PrimeKG edges and is capped to at most $L$ hops, reducing spurious, overly indirect chains. GPT-4o is then used only as a \emph{selector} over these candidates (it is instructed not to invent new nodes/edges and to preserve edge directionality and relation types; see Figure~\ref{fig:prompt_templates}). To improve stability, we use deterministic decoding (temperature $=0$) and a fixed JSON schema for path selection. In cases where multiple clinically plausible mechanisms exist, we retain up to $K_{\text{path}}$ paths and encourage diversity (non-redundant mechanisms) rather than a single ``best'' chain. We emphasize that the selected path set is intended as a compact \emph{scaffold} for CoT generation, not an exhaustive mechanistic catalog: alternative clinically valid paths may be omitted during pruning. 

\paragraph{CoT Generation with KG-Guided Reasoning.}
Utilizing the disease-specific KG reasoning paths $\mathcal{P}_d$ as guidance, we
distill reliable knowledge from the off-the-shelf KG into our CoT data. To achieve this, we
prompt the LLM to consider these paths together with KG-related evidence and the observed
feature nodes, and to elaborate them into medically grounded CoT explanations of the
prediction, represented as
\[
\begin{aligned}
C_{t,d}
&= \mathrm{LLM}\big(
    d,\;
    x_t^{+},\;
    \mathcal{R}_d^{+}(t),\;
    \mathcal{R}_d^{-}(t),\;
    \mathcal{P}_d,\;
    y_{t+1}^{(d)}
    \,\big|\,
    I_{\mathrm{gen}}
\big),\\[0.5ex]
\mathcal{R}_d^{+}(t)
&= \big\{ \hat e_{k,d} \in \mathcal{R}_d \;\big|\; x_t(\hat e_{k,d}) = 1 \big\}, 
\quad
\mathcal{R}_d^{-}(t)
= \mathcal{R}_d \setminus \mathcal{R}_d^{+}(t).
\end{aligned}
\]
Here $I_{\mathrm{gen}}$ is the CoT instruction prompt, and
$x_t \in \{0,1\}^{N}$ is the binary ICD\textendash9 feature vector for the index visit at time $t$,
with
$x_t^{+} = \{ c \mid x_t(c) = 1 \}$ denoting the set of ICD\textendash9 codes present at that visit.
The term $y_{t+1}^{(d)} \in \{0,1\}$ is the ground-truth label indicating whether disease $d$
is present at the next visit $t{+}1$. The sets $\mathcal{R}_d^{+}(t)$ and $\mathcal{R}_d^{-}(t)$
indicate which disease-relevant KG nodes mapped from PrimeKG are expressed and not
expressed at time $t$, respectively. The condition $x_t(\hat e_{k,d}) = 1$ means that the ICD\textendash9 code corresponding to KG node $\hat e_{k,d}$ is present in the feature vector
$x_t$.

\paragraph{Filtering.}
Each generated CoT $C_{t,d}$ is required to end with an explicit binary
conclusion $\hat{y}_{t+1}^{(d)} \in \{\text{Yes}, \text{No}\}$. We use this conclusion to
decide whether to keep the example: $C_{t,d}$ is kept if and only if the CoT-implied
label matches the ground-truth outcome $y_{t+1}^{(d)}$, i.e., $\hat{y}_{t+1}^{(d)} = y_{t+1}^{(d)}$.
The surviving set of $(x_t, d, C_{t,d}, y_{t+1}^{(d)})$ pairs forms a compact but
high-quality KG-grounded supervision corpus. We fine-tune lightweight open-weight
LLMs (e.g., LLaMA~\cite{grattafiori2024llama}, Gemma~\cite{team2024gemma}) on this dataset so that, at test time, they produce both predictions
and clinician-style reasoning traces.

\begin{figure}[h] 
  \centering
  \includegraphics[width=1\linewidth]{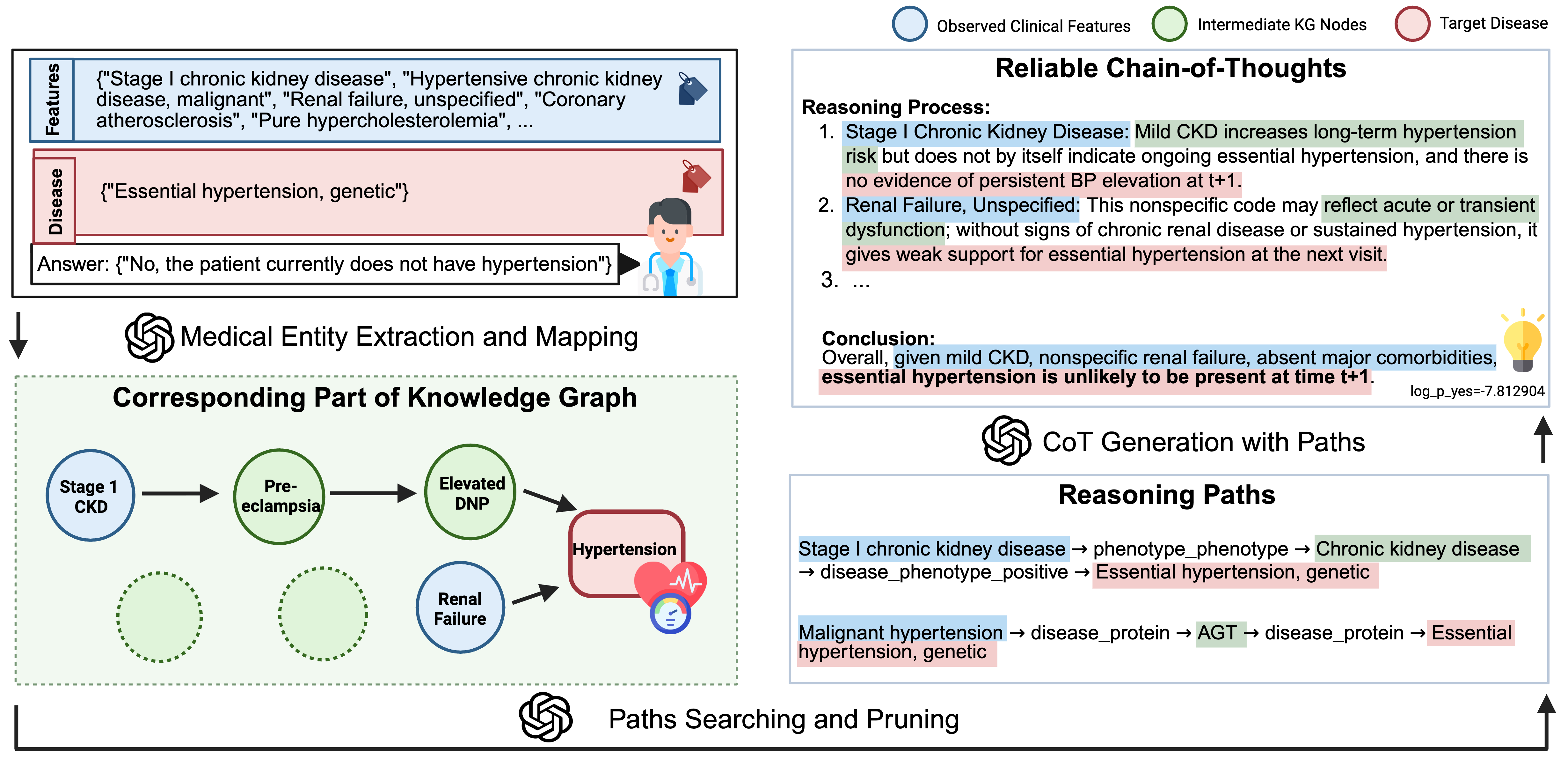}
  \caption{Schematic and example of KG-guided CoT generation and filtering. \textbf {Observed clinical features} are denoted in \textcolor{blueKey}{Blue}; 
\textbf{Intermediate KG nodes} are denoted in \textcolor{greenKey}{Green},
and the \textbf{Target disease} is denoted in \textcolor{redKey}{Red}.}
  \label{fig:my_figure}
\end{figure}

\section*{Results}
\refstepcounter{section}
\label{sec:results}

\paragraph{Dataset.} We work with longitudinal visits from the MIMIC-III critical care database.\cite{johnson2016mimic}
Following the setup of Harutyunyan et al.,\cite{harutyunyan2019multitask}
we focus on patients with more than one hospital visit and construct all pairs of adjacent
visits. For each such pair, the earlier visit serves as the index visit at time $t$ and the
later visit provides labels for the next visit at time $t{+}1$. This preprocessing yields 12{,}353 labeled index visits (i.e., visit pairs).

Each index visit at time $t$ is represented as a binary ICD-9 feature vector $x_t \in \{0,1\}^{N}$ with $N = 7{,}423$, where each dimension indicates the presence of a specific ICD-9 code during that visit. The original task is formulated as a multi-label prediction problem over 25 acute-care conditions derived from Clinical Classifications Software (CCS).\cite{harutyunyan2019multitask} In this work, we adopt the same 25 disease labels but evaluate on $|D|{=}10$ diseases that map cleanly to PrimeKG (Table~\ref{tab:mapped_diseases}). For each disease $d \in \mathcal{D}$ and index visit $t$, the prediction target is a binary label $y^{(d)}_{t+1} \in \{0,1\}$ indicating whether $d$ is present at the next visit $t{+}1$. Models take as input $(x_t, d)$ (and, for certain variants, KG evidence constructed as described in the Method section) and output a probability $\hat{p}^{(d)}_{t+1}$ and a binary verdict. e split 12{,}353 index visits by visit (10\% test) and subsample training sets of 400 and 1{,}000 visits for data-efficiency experiments, using the remainder for validation.

In addition to MIMIC-III, we evaluate zero-shot transfer on Project CRADLE (Emory Clinical Research Analytics Data Lake Environment), a de-identified EHR repository from Emory Healthcare.
Following prior work,\cite{xu2023hypehr} we focus on patients with type 2 diabetes and
formulate a binary prediction task: whether a patient will experience a 
\begin{wraptable}{r}{0.45\textwidth}
  \caption{The 10 MIMIC-III diseases mapped into PrimeKG and used as prediction targets.}
  \vspace{-0.1cm}
  \label{tab:mapped_diseases}
  \small
  \begin{tabular}{ll}
    \toprule
    \textbf{Mapped Disease Label} & \textbf{Category} \\
    \midrule
    Acute myocardial infarction & Cardiovascular \\
    Chronic kidney disease & Renal \\
    Chronic obstructive pulmonary disease & Respiratory \\
    Conduction disorders & Cardiovascular \\
    Coronary atherosclerosis & Cardiovascular \\
    Diabetes mellitus (no complication) & Metabolic \\
    Essential hypertension & Cardiovascular \\
    Gastrointestinal hemorrhage & Gastrointestinal \\
    Pneumonia & Infectious \\
    Shock & Critical \\
    \bottomrule
  \end{tabular}
  \vspace{-0.5cm}
\end{wraptable}
cardiovascular disease (CVD) event within one year of the initial diabetes diagnosis. CVD endpoints
include coronary heart disease, congestive heart failure, myocardial infarction, and stroke, identified via ICD-9/ICD-10 codes. After applying standard inclusion/exclusion criteria from the original CRADLE studies, the resulting cohort contains 36{,}611 patients with 12{,}724 binary features. 

\textbf{Evaluation Metrics.}
Since both the MIMIC-III and CRADLE cohorts exhibit substantial class imbalance, we
therefore report accuracy, area under the receiver operating characteristic curve (AUROC),
area under the precision--recall curve (AUPR), and macro-F1 as our primary evaluation
metrics. For accuracy and F1, we convert predicted probabilities to binary decisions using
a fixed threshold of 0.5. Table~\ref{tab:mapped_diseases} lists the 10 diseases used as prediction targets.

\paragraph{Experiment:} We conduct experiments under two training regimes: one using 400 index-visit cases and another using 1,000 index-visit cases. Unless otherwise noted, all models are evaluated using a fixed decision threshold of 0.5 on the predicted probabilities.

\begin{table*}[h]
\vspace{0.3cm}
\centering
\caption{Data-efficiency comparison on MIMIC-III visit-level disease prediction with 400 and 1000 index visits (10 diseases). Best metric in \textbf{bold}, second-best \underline{underlined}.}
\label{tab:mimic_all_results}
\setlength{\tabcolsep}{4pt}
\renewcommand{\arraystretch}{1.25}
\resizebox{\linewidth}{!}{%
\begin{tabular}{l*{2}{cccc}}
\toprule
\rowcolor{headgray}
\textbf{Model} &
\multicolumn{4}{c}{\textit{MIMIC-III (400 index visits)}} &
\multicolumn{4}{c}{\textit{MIMIC-III (1000 index visits)}} \\
\rowcolor{headgray}
\multicolumn{9}{c}{} \\[-1.6ex]
\cmidrule(r){2-5}
\cmidrule(l){6-9}
\rowcolor{headgray}
& Acc & AUROC & AUPR & F1 &
  Acc & AUROC & AUPR & F1 \\
\midrule

LLaMA3-8B (orig.) &
0.4709 & 0.5604 & 0.2773 & 0.3327 &
0.4709 & 0.5604 & 0.2773 & 0.3327 \\

\rowcolor{pastelgreen}
\textbf{LLaMA3-8B + KG-CoT (FT)} &
\textbf{0.8389} & \underline{0.6683} & \textbf{0.4049} & \textbf{0.4050} &
\textbf{0.8544} & \textbf{0.6995} & \textbf{0.4685} & \underline{0.4058} \\

Gemma-7B (orig.) &
0.7475 & 0.5152 & 0.2528 & 0.0169 &
0.7475 & 0.5152 & 0.2528 & 0.0169 \\

\rowcolor{pastelgreen}
\textbf{Gemma-7B + KG-CoT (FT)} &
\underline{0.8211} & 0.6648 & \underline{0.4002} & \underline{0.3976} &
\underline{0.8458} & 0.6609 & \underline{0.4363} & \textbf{0.4263} \\

\midrule

SGD  &
0.7277 $\pm$ 0.0251 & 0.6342 $\pm$ 0.0211 &
0.3595 $\pm$ 0.0235 & 0.1183 $\pm$ 0.0474 &
0.6886 $\pm$ 0.0243 & 0.6563 $\pm$ 0.0115 &
0.3840 $\pm$ 0.0114 & 0.1508 $\pm$ 0.0176 \\

\rowcolor{gray!15}
Logistic Regression  &
0.7503 $\pm$ 0.0096 & \textbf{0.6690} $\pm$ 0.0054 &
0.3970 $\pm$ 0.0067 & 0.0990 $\pm$ 0.0277 &
0.7280 $\pm$ 0.0158 & \underline{0.6874} $\pm$ 0.0105 &
0.4178 $\pm$ 0.0049 & 0.1238 $\pm$ 0.0138 \\

SVM  &
0.7385 $\pm$ 0.0000 & 0.6468 $\pm$ 0.0095 &
0.3824 $\pm$ 0.0032 & 0.1246 $\pm$ 0.0000 &
0.7391 $\pm$ 0.0011 & 0.6873 $\pm$ 0.0042 &
0.4256 $\pm$ 0.0054 & 0.1275 $\pm$ 0.0058 \\

\rowcolor{gray!15}
MLP  &
0.7334 $\pm$ 0.0105 & 0.6492 $\pm$ 0.0096 &
0.3803 $\pm$ 0.0145 & 0.1368 $\pm$ 0.0211 &
0.7029 $\pm$ 0.0299 & 0.6642 $\pm$ 0.0086 &
0.3911 $\pm$ 0.0069 & 0.1917 $\pm$ 0.0367 \\

RF  &
0.7999 $\pm$ 0.0030 & 0.6155 $\pm$ 0.0084 &
0.3687 $\pm$ 0.0053 & 0.1970 $\pm$ 0.0043 &
0.8107 $\pm$ 0.0024 & 0.6574 $\pm$ 0.0076 &
0.4024 $\pm$ 0.0049 & 0.2191 $\pm$ 0.0042 \\

\rowcolor{gray!15}
Naive Bayes &
0.5810 $\pm$ 0.0240 & 0.6010 $\pm$ 0.0162 &
0.3373 $\pm$ 0.0155 & 0.3672 $\pm$ 0.0090 &
0.5394 $\pm$ 0.0185 & 0.6224 $\pm$ 0.0039 &
0.3581 $\pm$ 0.0064 & 0.3836 $\pm$ 0.0036 \\

XGBoost  &
0.8018 $\pm$ 0.0037 & 0.6476 $\pm$ 0.0053 &
0.3938 $\pm$ 0.0085 & 0.2090 $\pm$ 0.0058 &
0.8159 $\pm$ 0.0018 & 0.6799 $\pm$ 0.0095 &
0.4309 $\pm$ 0.0060 & 0.3065 $\pm$ 0.0058 \\

\bottomrule
\end{tabular}%
}
\end{table*}
As shown in Table~2, KG-guided CoT fine-tuning substantially boosts the performance of both LLaMA-3.1-8B and Gemma-7B in the low-data MIMIC-III setting. With only 400 labeled index visits data, the KG-CoT--tuned LLaMA model attains the best accuracy and macro-F$_1$ and the highest macro-AUPR across all methods, while matching the AUROC of the strongest logistic regression baseline ($\approx 0.67$). Gemma-7B shows a similar trend: KG-CoT supervision raises its macro-AUPR from 0.25 to 0.40 and macro-F$_1$ from essentially zero to $\approx 0.40$, bringing it in line with the tuned LLaMA variant. When the training set is increased to 1{,}000 cases, both KG-guided models improve further, with LLaMA-3.1 achieving the top accuracy, AUROC ($\approx 0.70$), and macro-AUPR ($\approx 0.47$), and Gemma-7B achieving the highest macro-F$_1$, consistently outperforming classical baselines such as XGBoost and Random Forest. These results indicate that KG-anchored CoT supervision enables small open-weight LLMs to learn data-efficient, clinically grounded prognostic reasoning that is competitive with or superior to traditional EHR prediction models.

\newpage

\begin{wrapfigure}[20]{r}{0.45\textwidth}
\vspace{0.3cm}
\centering

% -------- Table 3 --------
\captionof{table}{Ablation on MIMIC-III visit-level disease prediction with 1000 patient index visits. Best metric in \textbf{bold}, second-best \underline{underlined}.}
\vspace{-0.2cm}
\label{tab:ablation_1000_visits}
\setlength{\tabcolsep}{4pt}
\renewcommand{\arraystretch}{1.4}
\scalebox{0.75}{
\begin{tabular}{lcccc}
\toprule
\rowcolor{headgray}
\textbf{Setting} & Acc & AUROC & AUPR & F1 \\
\midrule
LLaMA3-8B + KG (unfiltered) & 0.8485 & 0.6589 & 0.4473 & 0.4023 \\
LLaMA3-8B (no KG, filtered) & \textbf{0.8613} & \underline{0.6640} & \underline{0.4637} & \textbf{0.4357} \\
\rowcolor{pastelgreen}
\textbf{LLaMA3-8B + KG} & \underline{0.8544} & \textbf{0.6995} & \textbf{0.4685} & \underline{0.4058} \\
\bottomrule
\end{tabular}}

\vspace{0.7cm} % space between the two right-side tables

% -------- Table 4 --------
\captionof{table}{Zero-shot transfer on the CRADLE cohort. Best metric in \textbf{bold}, second-best \underline{underlined}.}
\label{tab:cradle_final}
\vspace{-6pt}
\setlength{\tabcolsep}{4pt}
\renewcommand{\arraystretch}{1.4}
\scalebox{0.75}{
\begin{tabular}{lcccc}
\toprule
\rowcolor{headgray}
\textbf{Model} & Acc & AUROC & AUPR & F1 \\
\midrule
Llama3-8b-it (orig.) & 0.4026 & 0.5170 & 0.2241 & 0.3446 \\
\rowcolor{pastelgreen}
\textbf{Llama3-8b-it-400\_data (FT)} & 0.7508 & \underline{0.5176} & \textbf{0.2512} & \textbf{0.5143} \\
\rowcolor{pastelgreen}
\textbf{Llama3-8b-it-1000\_data (FT)} & \textbf{0.7686} & \textbf{0.5348} & \underline{0.2488} & 0.4567 \\
\midrule
Gemma-7b (orig.) & 0.5102 & 0.5052 & 0.2198 & 0.3064 \\
\rowcolor{pastelgreen}
\textbf{Gemma-7b-400\_data (FT)} & 0.7217 & 0.5023 & 0.2209 & \underline{0.4905} \\
\rowcolor{pastelgreen}
\textbf{Gemma-7b-1000\_data (FT)} & \underline{0.7654} & 0.4845 & 0.2121 & 0.4763 \\
\bottomrule
\end{tabular}}

\vspace{-0.3cm}
\end{wrapfigure}\ignorespaces

\noindent To isolate KG guidance and filtering effects, we compare three variants at 1{,}000 index visits in Table~3. Removing KG guidance or using unfiltered CoT traces reduces AUROC/AUPR, suggesting that ungrounded or noisy rationales degrade ranking performance. Overall, KG-grounded and curated CoT supervision is key to strong AUROC/AUPR in this regime. 

\noindent To assess cross-cohort robustness, we evaluate the MIMIC-trained models zero-shot on CRADLE (36{,}611 patients; 12{,}724 binary features) to predict cardiovascular disease within one year after type~2 diabetes diagnosis, without re-training or calibration. MIMIC-trained models transfer to the CRADLE cohort to forecast cardiovascular disease within one year after the diagnosis of type~2 diabetes. For the LLaMA-3.1 backbone, fine-tuning on 400--1{,}000 MIMIC-III cases increases accuracy from 0.40 to 0.75--0.77 and macro-F$_1$ from 0.34 to 0.46--0.51, while modestly raising AUROC (to $\approx 0.52$–0.53) and AUPR (to $\approx 0.25$). Gemma-7B exhibits a comparable pattern: accuracy rises from 0.51 to 0.72--0.77 and macro-F$_1$ from 0.31 to about 0.48, again with modest changes in AUROC and AUPR. These results suggest that the reasoning learned from KG-guided CoT supervision on MIMIC-III generalize across cohorts and spaces, enabling zero-shot transfer of backbones to out-of-distribution EHR data.
% \begin{table}[t]
% \centering
% \caption{Per-disease performance on MIMIC-III (LLaMA-8B, 1000 index visits; 10 diseases).}
% \label{tab:mimic_per_disease_llama8b_1000}
% \setlength{\tabcolsep}{4pt}
% \renewcommand{\arraystretch}{1.15}
% \begin{tabular}{lcccc}
% \toprule
% \textbf{Disease} & \textbf{Acc} & \textbf{AUROC} & \textbf{AUPR} & \textbf{F1} \\
% \midrule
% Acute myocardial infarction & 0.9328 & 0.5332 & 0.0769 & 0.0000 \\
% Chronic kidney disease & 0.8753 & 0.8027 & 0.6498 & 0.7220 \\
% Chronic obstructive pulmonary disease & 0.8664 & 0.6991 & 0.4990 & 0.5946 \\
% Conduction disorders & 0.9619 & 0.7110 & 0.1855 & 0.3562 \\
% Coronary atherosclerosis and related & 0.8737 & 0.8348 & 0.8025 & 0.8138 \\
% Diabetes mellitus without complication & 0.8105 & 0.7479 & 0.7568 & 0.7234 \\
% Essential hypertension & 0.8057 & 0.7849 & 0.8377 & 0.7595 \\
% Gastrointestinal hemorrhage & 0.7142 & 0.4775 & 0.3008 & 0.0536 \\
% Pneumonia & 0.7684 & 0.5147 & 0.2494 & 0.0000 \\
% Shock & 0.8761 & 0.4832 & 0.1150 & 0.0000 \\
% \bottomrule
% \end{tabular}
% \end{table}
\paragraph{Human Evaluation.}
To evaluate whether KG-guided CoT improves explanation quality, we conducted a blinded human study comparing our KG-anchored LLaMA-3.1 model with its untuned baseline.
For each of 115 randomly sampled cases, clinicians were shown an input summary, the ground-truth outcome, and two anonymized model outputs (A and B), each containing a prediction and its reasoning trace (one KG-guided, one baseline).
Without knowing which model produced which trace, clinicians made a pairwise preference judgment (A vs.\ B) along three complementary dimensions:
\vspace{-0.3cm}

\begin{enumerate}
\item \textbf{Clarity \& Coherence (presentation):} he trace is easy to follow, logically ordered, and avoids jumps, circular reasoning, or contradictions.
\item \textbf{Coverage \& Relevance (evidence selection):} the trace surfaces key patient-specific factors for the target disease, stays clinically focused, and avoids generic or distracting content.
\item \textbf{Correctness \& Soundness (clinical validity):} clinical claims and timelines are medically plausible and consistent with the provided features/outcome (no fabricated details or inappropriate causal leaps), and the conclusion follows from the stated evidence.
\end{enumerate}

\vspace{-0.3cm}
These dimensions are intentionally distinct: \emph{clarity} evaluates how the reasoning is communicated, \emph{coverage} evaluates what evidence is selected and emphasized, and \emph{correctness} evaluates whether the medical content is accurate and logically warranted. The annotation was performed by two human experts from different medical schools: an MD professor and an MD/PhD student in biomedical informatics.
In total, 115 unique cases were annotated, with the MD professor reviewing 65 cases and the MD/PhD student reviewing 50 cases.

\textbf{Inter-rater consistency.}
To quantify consistency across adjudicators, we additionally double-annotated a random subset of $n_{\text{overlap}}$ of 55 cases (both clinicians independently rated the same cases across all three dimensions) and computed Cohen's $\kappa$ on the binary preference decision (KG-guided vs.\ baseline).
Agreement was $\kappa_{\text{clar}}$ for \emph{Clarity \& Coherence}, $\kappa_{\text{cov}}$ for \emph{Coverage \& Relevance}, and $\kappa_{\text{corr}}$ for \emph{Correctness \& Soundness}.

% \begin{figure*}[h]
%   \centering
%   \includegraphics[width=.8\textwidth]{figs/pi-chart.png}
%   \caption{Clinician preference for KG-guided explanations over the untuned baseline across three dimensions. Each donut shows the percentage of cases
%   in which the KG-guided output was preferred; the baseline share is the complement.}
%   \label{fig:human_eval_pies}
% \end{figure*}

\begin{wrapfigure}{r}{0.5\textwidth}
  \vspace{-0.5\baselineskip} % tweak if it sits too low/high
  \centering
  \includegraphics[width=\linewidth]{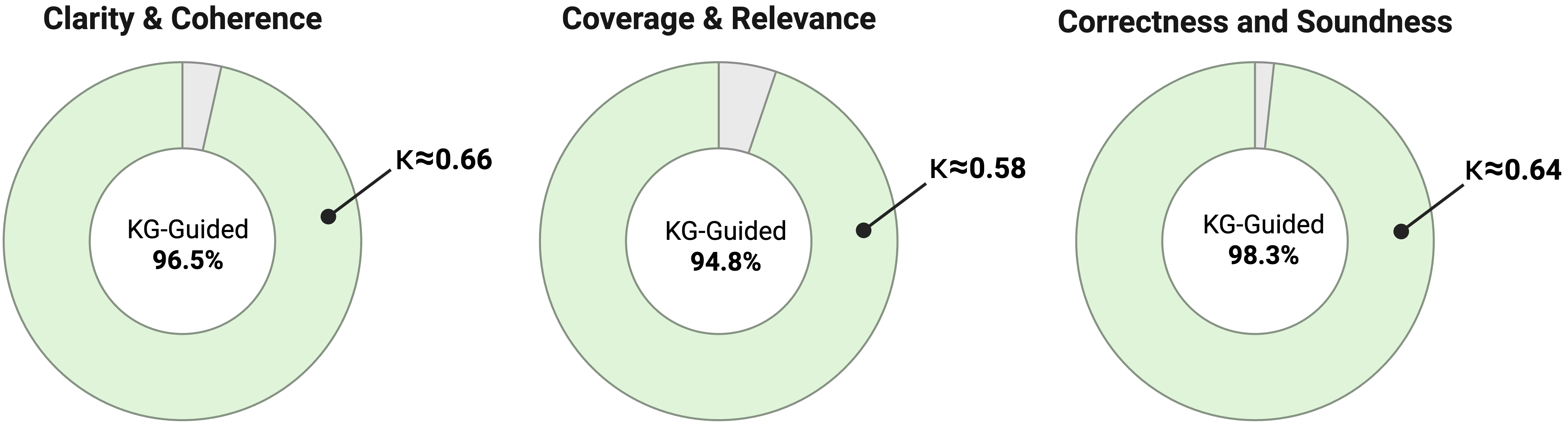}
  \caption{Clinician preference for KG-guided explanations vs. the untuned baseline across three dimensions (donuts show \% preferred; remainder is baseline).}
  \label{fig:human_eval_pies}
  \vspace{1pt} % tweak to control gap after figure
\end{wrapfigure}

On the 115 annotated cases, both clinicians preferred the KG-guided model (Model~A) over the untuned baseline (Model~B): 96.5\% for \emph{Clarity \& Coherence}, 94.8\% for \emph{Coverage \& Relevance}, and 98.3\% for \emph{Correctness \& Soundness}. Qualitatively, clinicians noted that KG-guided CoT traces more often (i) organized evidence temporally, (ii) highlighted clinically meaningful risk factors over superficial code patterns, and (iii) avoided unsupported leaps or hallucinated details. By contrast, the untuned baseline was described as more ``trigger-happy'' (more likely to conclude \emph{YES} from weak signals). While both models are sensitive to poor or incomplete inputs, the baseline was more prone to repetitive differentials and illogical reasoning, especially on short or noisy inputs.
\section*{Discussion}
\label{sec:discussion}

We introduce a KG-guided chain-of-thought (CoT) framework that uses PrimeKG structure to scaffold LLM reasoning over structured EHRs.
By mapping ICD\textendash9 codes to PrimeKG,\cite{chandak2023primekg} mining disease-specific relevance sets and short paths, and applying label-consistency filtering, we build a compact supervision corpus for data-efficient fine-tuning of lightweight LLMs.

\paragraph{Finding 1: KG-anchored CoT improves low-data learning, especially under imbalance.}
Across both 400 and 1{,}000 labeled index visits (Table~\ref{tab:mimic_all_results}), KG-guided CoT substantially improves LLaMA-3.1-8B and Gemma-7B and is competitive with strong classical baselines.
Gains are largest on macro-AUPR (e.g., $\approx 0.25 \rightarrow 0.40$ for Gemma at 400), suggesting that KG-anchored supervision encourages disease-relevant evidence over superficial code co-occurrence.

\paragraph{Finding 2: Filtering and KG evidence play complementary roles.}
In the ablation (Table~\ref{tab:ablation_1000_visits}), filtering alone yields the best accuracy/F$_1$, while adding KG evidence yields the best AUROC (up to $\approx 0.70$). Practically, filtering reduces noisy supervision and spurious positives, whereas KG structure strengthens ranking; threshold calibration can further improve accuracy/F$_1$. Future work can further improve threshold-dependent metrics (accuracy/F$_1$) by calibrating decision thresholds per disease or cohort.

\paragraph{Finding 3: Zero-shot transfer mainly shifts the operating point, motivating calibration.}
On CRADLE (Table~\ref{tab:cradle_final}), fine-tuned models boost accuracy (0.72--0.77) with only modest AUROC/AUPR changes, indicating a changed decision rule under a fixed threshold rather than large ranking gains.
Deployment evaluations should therefore include calibration and threshold selection, especially across cohorts with different prevalences and coding practices.

\paragraph{Finding 4: Clinicians preferred KG-guided explanations.}
Clinicians preferred KG-guided traces on clarity/coherence, coverage/relevance, and correctness/soundness for most cases.
They cited better temporal organization, more clinically meaningful risk factors, and fewer unsupported leaps, suggesting KG evidence as an effective planning scaffold for clinician-facing rationales.

\paragraph{Limitations and future work.}
Our approach depends on the coverage and quality of PrimeKG, so gaps or biases in the KG can propagate into both supervision and explanations.
The ICD to KG mapping and the relevance/path selection steps rely on GPT-4o and heuristic constraints. 
While we constrain candidate paths using shortest-path enumeration and use the LLM only as a selector, the pruned set is not guaranteed to be exhaustive: alternative clinically valid paths may be removed in favor of a compact scaffold.
Methodologically, we also adopt a single-visit formulation ($x_t$) for next-visit prediction, which simplifies evaluation but does not fully exploit longitudinal history.
Future work can extend the framework to richer EHR views (notes, medications, labs), incorporate multi-visit context, and integrate clinician-in-the-loop validation for KG evidence selection and pruning.
\section*{Conclusion}
\label{sec:conclusion}
In this paper, we introduced a KG-guided chain-of-thought framework that couples biomedical structure from PrimeKG with LLM-based reasoning over structured EHR data. By using KG paths as scaffolds for CoT generation and label-consistent filtering, we constructed a compact supervision corpus that enables small open-weight LLMs to achieve competitive or superior visit-level disease prediction on MIMIC-III under strict data constraints, while also producing explicit, clinically grounded explanations. Zero-shot transfer to the CRADLE cohort suggests that the learned decision boundaries generalize across cohorts and feature spaces, and blinded clinician review indicates a consistent preference for KG-guided CoT over untuned LLM explanations. Taken together, these results highlight KG-anchored CoT supervision as a promising direction for building clinical decision-support systems in which LLMs not only predict risk but also articulate transparent, patient-level reasoning aligned with biomedical knowledge and clinician expectations.
\section*{Acknowledgements}
\vspace{-0.2cm}
This research was partially supported by the internal funds and GPU servers provided by the Computer Science Department of Emory University, the US National Science Foundation under Award Numbers 2442172, 2312502, 2319449, and the US National Institutes of Health under Award Numbers K25DK135913, RF1NS139325, R01DK143456 and U18DP006922.
\bibliographystyle{unsrt}
\bibliography{amia}

\paragraph{Implementation Details.}
We fine-tuned LLaMA-3.1-8B-Instruct and Gemma-7B-Instruct on a single NVIDIA H200 (512\,GB host memory; Emory HPC); full hyperparameters and training configuration are provided in our GitHub repository. We selected the checkpoint with the best validation macro-AUPR. KG-guided CoT traces were generated with GPT-4o; CoT data generation creation cost \mbox{$\approx\textdollar 100$}. Prompt templates are provided below.

\begin{figure*}[h]
  \centering
  \begin{minipage}{0.48\linewidth}
    \centering
    \includegraphics[width=\linewidth]{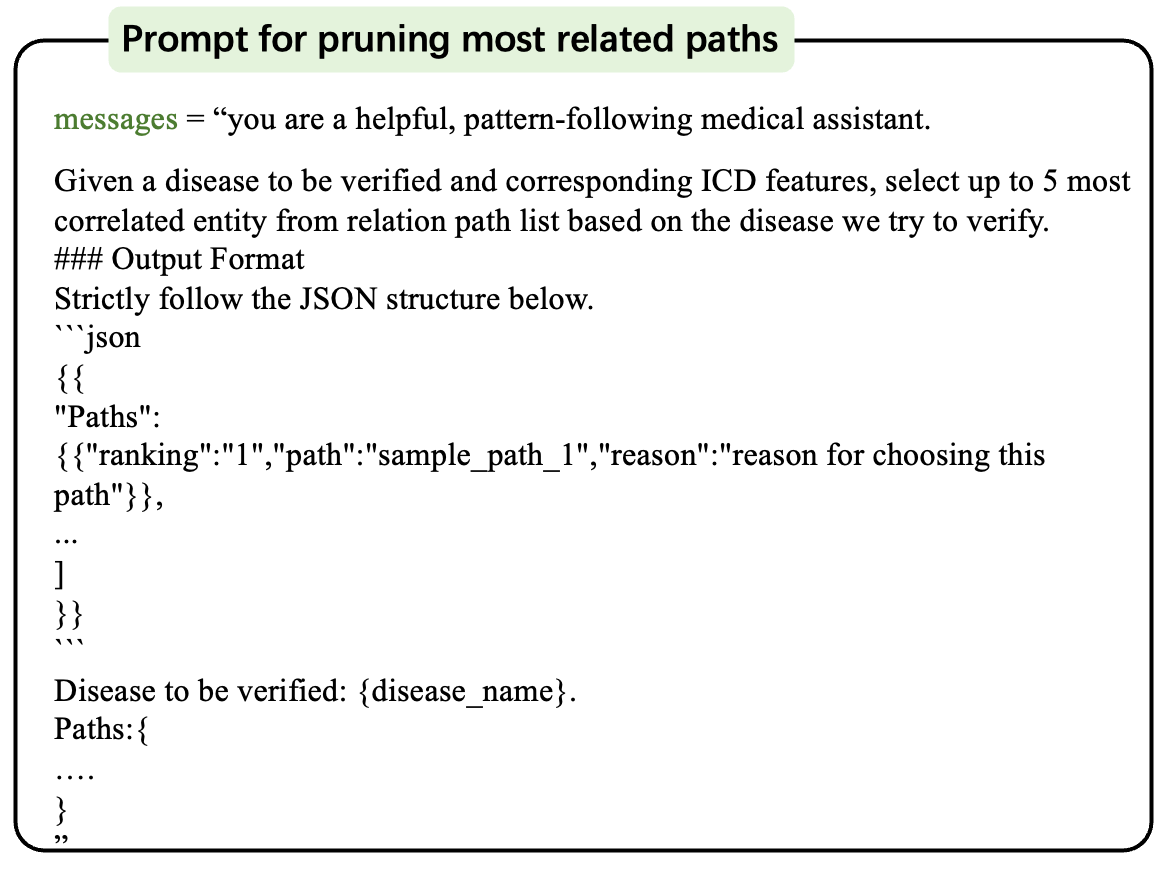}
  \end{minipage}
  \hfill
  \begin{minipage}{0.48\linewidth}
    \centering
    \includegraphics[width=\linewidth]{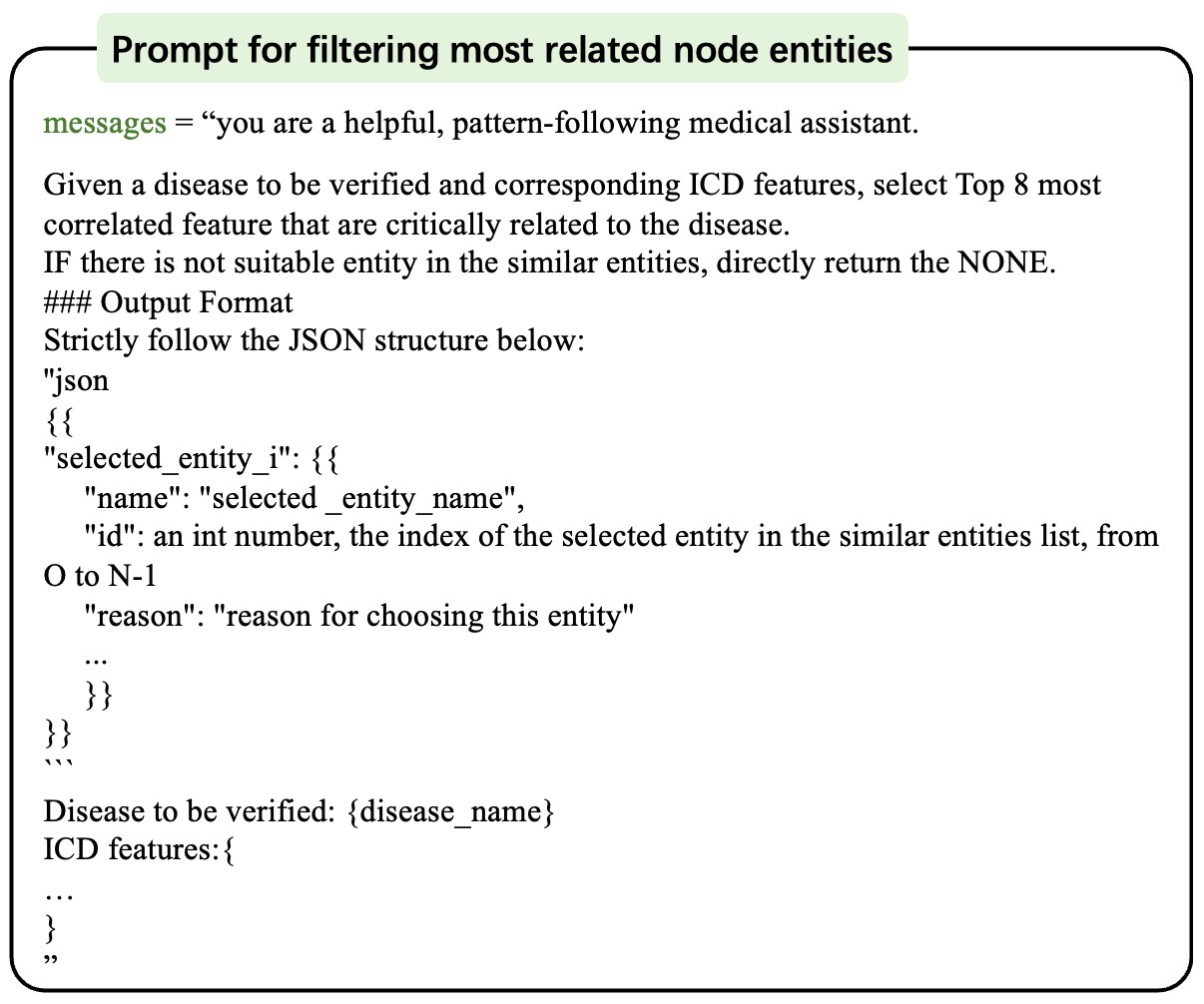}
  \end{minipage}

  \vspace{0.5em}

  \includegraphics[width=0.8\linewidth]{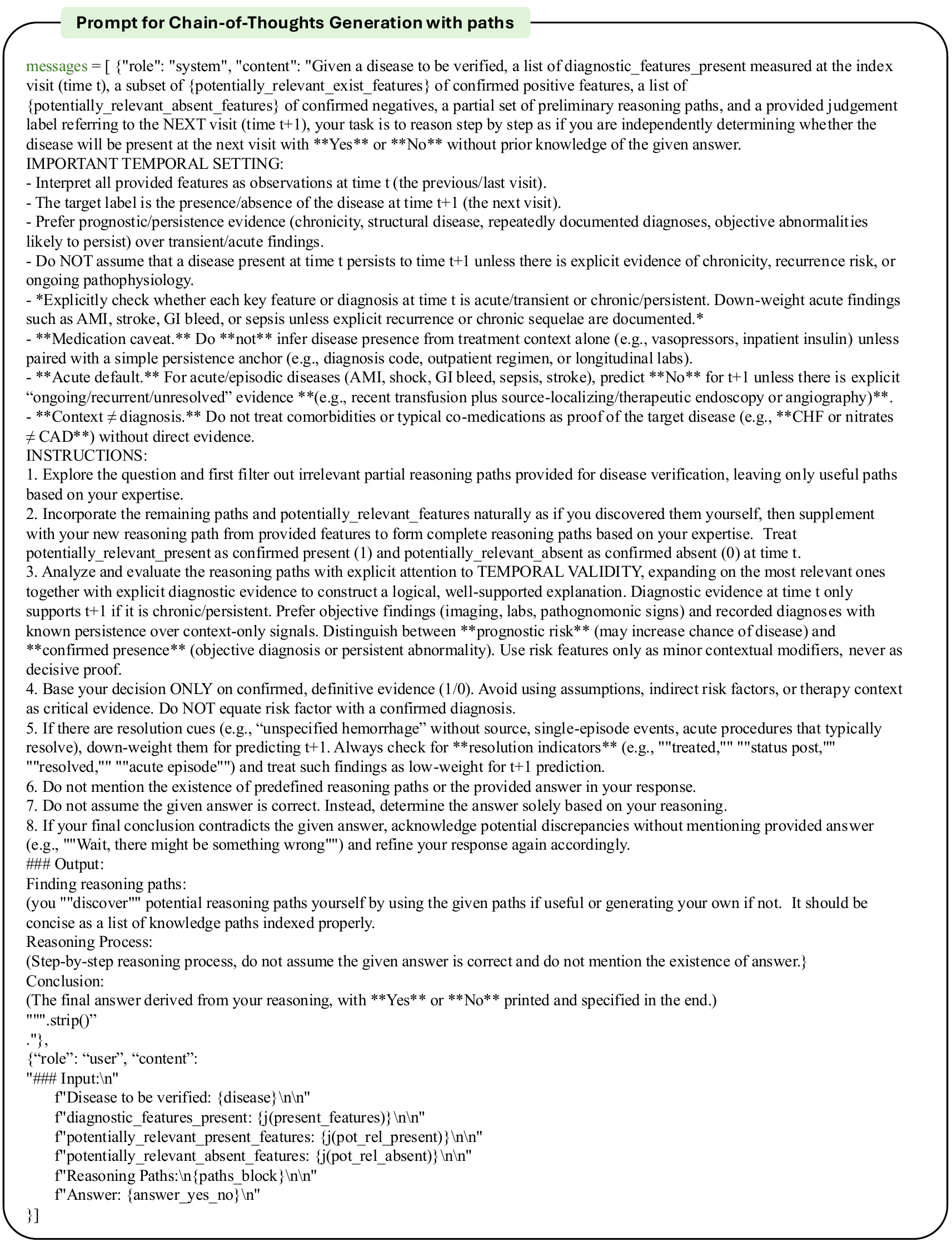}
  \vspace{-0.2cm}
  \caption{Prompt templates used in our KG-guided CoT pipeline:
  (top-left) disease-relevant node selection,
  (top-right) KG path pruning and selection,
  (bottom) CoT generation conditioned on KG evidence and visit features.}
  \label{fig:prompt_templates}
\end{figure*}

\end{document}